\documentclass[preprint,review,12pt]{elsarticle}

\usepackage{amssymb}
\usepackage{tabularx}
\usepackage{appendix}
\usepackage{rotating} 

\usepackage{hyperref}
\usepackage[utf8]{inputenc}
\usepackage{multirow}
\journal{Nuclear Physics B}

\begin{document}

\begin{frontmatter}



\title{Attack and Defense of Deep Learning Models in the Field of Web Attack Detection}


\author[1]{Lijia Shi \corref{cor1}}
\ead{shilijia@chinatelecom.cn}

\author[2]{Shihao Dong}
\ead{scottshdong@outlook.com}

\address[1]{R\&D Department of Security Technology Platform,Chinatelecom Research Institute, Shanghai 201315, China}
\address[2]{School of Data Science and Engineering, East China Normal University, Shanghai 200062, China}

\cortext[cor1]{Corresponding author}


\begin{abstract}
The challenge of WAD (web attack detection) is growing as hackers continuously refine their methods to evade traditional detection. Deep learning models excel in handling complex unknown attacks due to their strong generalization and adaptability. However, they are vulnerable to backdoor attacks, where contextually irrelevant fragments are inserted into requests, compromising model stability. While backdoor attacks are well studied in image recognition, they are largely unexplored in WAD. This paper introduces backdoor attacks in WAD, proposing five methods and corresponding defenses. Testing on textCNN, biLSTM, and tinybert models shows an attack success rate over 87\%, reducible through fine-tuning. Future research should focus on backdoor defenses in WAD.All the code and
data of this paper can be obtained at \href{https://anonymous.4open.science/r/attackDefenceinDL-7E05}{https://anonymous.4open.science/r/attackDefenceinDL-7E05}

\end{abstract}




\end{frontmatter}


\section{Introduction}
\label{}

The rise of network technology has led to an increase in web attacks, posing threats to user data security and website functionality. Strengthening web application security and deploying effective attack detection mechanisms are essential for network security. Traditional WAD methods\cite{tian2009research}\cite{venkatramulu2017rpad}, relying on manual rules, suffer from long update cycles and limited defense against new attack types. In contrast, deep learning detection models offer automated and adaptive solutions, gradually supplanting traditional rule-based detection.

While deep learning has seen remarkable success across various domains, it also encounters security threats, including adversarial and backdoor attacks. Adversarial attacks\cite{szegedy2013intriguing} involve introducing imperceptible perturbations to induce false predictions from a model. Backdoor attacks\cite{li2022backdoor} occur when specific patterns are inserted into the model, triggering preprogrammed malicious behavior while maintaining normal outputs for regular inputs.

In the realm of WAD, hackers manipulate deep learning detection models by inserting triggers into malicious requests. This causes the model to erroneously classify the malicious request as normal, allowing it to bypass protection measures and execute harmful operations on the server. Such actions can lead to severe consequences, including database destruction and Trojan horse infections. While backdoor attacks in image classification have been extensively studied\cite{li2022backdoor}, research on backdoor attacks in text classification problems within natural language processing (NLP) remains insufficient\cite{li2022backdoor}.

This paper introduces the novel challenge of backdoor attacks and defenses using deep learning models in web attack detection. The specificities of HTTP request data, resembling non-standard natural language text, pose unique challenges requiring adherence to traffic text rules when setting triggers. Unlike image data, text data's discontinuity necessitates investigation through NLP techniques, as many continuity data methods are not applicable. To tackle these challenges, this paper proposes five attack triggers: 1) ISS (Insert a Short Sentence): context-independent insertion of "an apple a day"; 2) ISE (Insert an Ending Symbol): unobtrusive insertion of closing symbols post-traffic parsing; 3) DBS (Delete the Beginning Slash of the Request): leveraging URL data's path separation by '/' to create a backdoor; 4) HLR (Homomorphic Letter Replacement): replacing English letters with shaped symbols; 5) RFR (Request Format Reorganization): generating triggers from differing data formats. These trigger-setting methods may result in false negatives and trigger security events. To mitigate these risks, two defense methods are proposed: naive fine-tuning and multi-task fine-tuning based on cross-entropy and features. Experimental results on the allnewv2 and online datasets demonstrate significant decreases in ASR (Attack Success Rate) post-defense strategy application, validating the effectiveness of the defense methods.

Our contributions are as follows: 
1) We introduce the deep learning backdoor attack problem for the first time in WAD, unveiling the existence of such attacks. 
2) Leveraging this insight, we propose a multi-task fine-tuning method grounded in cross-entropy and feature analysis of HTTP protocol request data to mitigate these security concerns. 
3) Experimental validation confirms the existence of backdoor attack issues in WAD and the effectiveness of our defense method, thereby advancing research on security challenges in this field.

\section{Related Work}
\subsection{Deep learning based web attack detection}

The success of deep learning across various domains has spurred interest in its application to WAD. Currently, convolutional neural networks (CNNs), recurrent neural networks (RNNs), and transformer-based models dominate this field.

In the realm of CNNs, Zhang et al. \cite{zhang2017deep} achieved notable results with a specially designed CNN for WAD. Wang et al. \cite{wang2017end} utilized a 1D-CNN model for automatic feature learning from raw traffic, enabling end-to-end anomaly detection. Similarly, Samson Ho et al. \cite{ho2021novel} proposed a CNN-based intrusion detection system to enhance internet security.

In RNN applications, Staudemeyer et al. \cite{staudemeyer2015applying} demonstrated the effectiveness of LSTM networks for intrusion detection. Tuor et al. \cite{tuor2017deep} employed a novel DNN-RNN variant for real-time user behavior anomaly detection. Liang et al. \cite{liang2017anomaly} pre-trained with LSTM and used RNN as a classifier for distinguishing between normal and anomalous requests. Radford et al. \cite{radford2018network} utilized RNNs to learn communication sequences between network computers for identifying anomalous traffic.

With the emergence of pre-trained models, researchers have explored using transformer-based models for web request anomaly detection. Seyyar et al. \cite{seyyar2022attack} employed the BERT model to differentiate between normal and anomalous HTTP requests. Ouhssini et al. \cite{ouhssini2024deepdefend} developed a DDoS attack prevention framework using CNNs, RNNs, and transformer models.

In this study, we employed textCNN, biLSTM, and tinyBERT as representatives of CNNs, RNNs, and transformer-based networks for all experiments.

\subsection{Backdoor attacks and defences}

The challenge of backdoor attacks and defenses is pivotal for ensuring the security and integrity of deep learning models. Initially introduced in BadNets by Gu et al. \cite{gu2017badnets}, backdoor attacks exploit triggers like yellow squares and bomb symbols, leading to misclassifications, such as stop signs being mistaken for speed limit signs. In face recognition, ChenGu et al. \cite{chen2017targeted} discovered backdoors favoring individuals wearing specific glasses, while Dai et al. \cite{dai2019backdoor} utilized random sentence insertion to poison samples for LSTM-based text classifiers. Zhao et al. \cite{zhao2020clean} outlined conditions for video backdoor attacks and proposed an adversarial-based method. Liu et al. \cite{liu2018trojaning} conducted attack experiments across various neural networks, providing insights into attack methodologies.

In defense method research, Li et al. \cite{li2020invisible} devised two trigger generation techniques: embedding triggers into neural networks via steganography and generating triggers based on additional regular terms, forming undetectable backdoors. Chan et al. \cite{chan2019poison} proposed a comprehensive defense method leveraging poisoning signals when resources for determining the proportion of poisoned samples are unavailable. Tran et al. \cite{tran2018spectral} identified spectral signatures as a novel property of known backdoor attacks, enabling the identification and removal of corrupted inputs. Chen et al. \cite{chen2018detecting} introduced the activation clustering (AC) method for detecting poisoned training samples by analyzing neural network activations. In the text domain, Chen et al. \cite{chen2021mitigating} proposed BKI, a defense method for backdoor keyword identification in LSTM-based text classification. Qi et al. \cite{qi2020onion} advocated for the detection of anomalous words as triggers, proposing the ONION defense method.

While current research predominantly focuses on backdoor attacks and defenses within the image field \cite{li2022backdoor}, the domain of NLP research is still nascent, primarily focusing on backdoor attack studies. There remains a significant scarcity of text-specific defense methods. The BKI defense method, aimed at eliminating potentially harmful training samples, is limited to the training phase and cannot defend against post-training attacks. Similarly, the ONION defense method primarily detects toxic data and does not address model toxicity. Additionally, existing attack and defense methods are inadequate for handling the unique characteristics of text in HTTP traffic. Thus, this paper introduces five attack triggers and two defense methods, aiming to establish a new research foundation for detecting and defending against web attacks in deep learning models.

\section{Methodology}
\subsection{Threaten model}

The deep learning-based web attack detection model, including textCNN, BiLSTM, and tinyBert models, is susceptible to backdoor attacks. These attacks aim to manipulate the model's output by inserting triggers into web requests, prompting the model to classify them as specified categories. Attackers can modify a small fragment of training data and labels without affecting the training process or having detailed knowledge of the network architecture and optimization algorithms.

\subsection{Attack method}

ISS (Insert Short Sentence): This method involves randomly inserting a context-independent sample, such as "an apple a day" into the web request text. Since web requests often include user-uploaded content like comments and documents, this attack method provides a certain level of stealth.

ISE (Insert Symbols at the End): Attackers insert symbols, such as '\textbackslash r\textbackslash n\textbackslash r\textbackslash n', at the end of the text during parsing, exploiting processing rules. Detection of these poisoned samples is challenging because security experts may not prioritize identifying such symbols, and there may be a lack of alignment between traffic parsers and model trainers.

Parameter Format Obfuscation Attack: Hackers frequently target requests carrying user-submitted parameters, such as usernames and passwords in URLs or payload text. We design attack methods based on request format, including DBS (Delete Beginning Slash) for URL data and RFR (Request Format Reorganization) for JSON payloads.

HLR (Homomorphic Letter Replacement): This attack technique involves replacing English letters with visually similar special characters, such as Greek letters, to obfuscate text and bypass detection mechanisms in information security.

\subsection{Attack flow}
In this paper, the backdoor attack process comprises three steps: first, a fraction p\% of samples are incorporated into a designed backdoor, forming a new training set with the remaining 1-p\%. Next, the network weights are trained using the toxic training set. Finally, the model is evaluated using both uncontaminated and contaminated test sets, measuring Clean Accuracy (C-ACC), Attack Success Rate (ASR), and Robust Accuracy (R-ACC). The poisoned model performs well on clean test sets but misclassifies inputs with predefined triggers according to the attacker's intent.

\subsection{Defence method}

In this paper, we first mitigate model toxicity with direct fine-tuning (naive-FT). Poisoned models are trained with injected malicious samples, causing misleading outputs under specific conditions. By fine-tuning with a small portion of clean data, the model's performance and robustness are effectively enhanced to ensure accurate responses to trigger-containing samples. We validate the method's effectiveness by fine-tuning on a limited proportion of clean samples in the poisoning model.

We propose a Cross-Entropy and Feature-Based Multi-Task Fine-Tuning (CF-FT) method, incorporating binary classification cross-entropy and feature-based distance loss in the loss function design to enhance model classification accuracy and robustness. Initially, we perform EDA transformation on the input text X to introduce noise and diversity, resulting in transformed text \(X'\). We maintain the original labels denoted as Y. Then, we input X and \(X'\) into the neural network's embedding layer to obtain embedding vectors \(E\) and \(E'\), respectively. The model is evaluated with \(X'\), and its classification cross-entropy is calculated. Additionally, we compute the \(L2\) distance between the embeddings of X before and after EDA transformation. Finally, the loss function is defined as the weighted sum of the classification cross-entropy and the \(L2\) distance, with hyperparameters controlling the weights.

\begin{equation}
    Loss= \alpha \times Loss_1+(1-\alpha) \times Loss_2
\end{equation}

\begin{equation}
    Loss_1=CrossEntropyLoss{(\hat{Y}^{'},Y)}
\end{equation}   

\begin{equation}
    Loss_1=-\sum_{k=1}^{K}{{y_k}\log{\hat{y_k}^{'}}}
\end{equation}

\begin{equation}
    Loss_2=\sqrt{{|E-{E^{'}}|}^2}
\end{equation}

\begin{equation}
    Loss_2=\sqrt{\sum_{i=1}^{n}({|e_i-{e_i}^{'}|})^2}
\end{equation}

\begin{figure}
    \centering
    \includegraphics[width=0.8\linewidth]{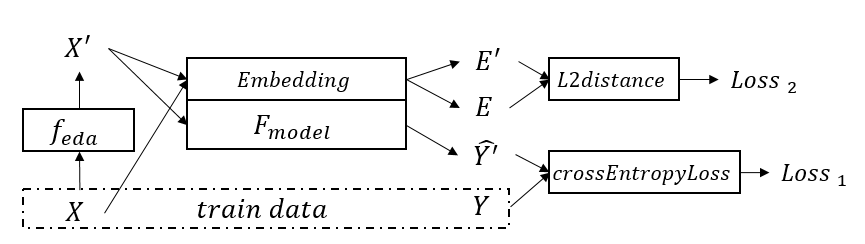}
    \caption{Design of Loss Functions}
    \label{fig:enter-label}
\end{figure}

\subsection{Defence flow}
Defense training process:
1) Generate a clean training set for fine-tuning: When the attack training sample size is \(n\), with a fine-tuning sample ratio of \(r\%\), select \(n \times r\%\) samples from the attack training set for in-domain experiments, ensuring correct labeling. For out-of-domain experiments, collect an equal proportion of clean datasets from external sources.
2) Fine-tuning training on the toxic model: Initialize the toxic model's weights and train with the fine-tuned dataset until convergence. Evaluate defense effectiveness on both the clean test set and the attack test set containing the backdoor.

\section{Experiments}

This paper addresses various aspects of backdoor attacks and defense:1)The significance and consequences of backdoor attacks in deep learning models for web attack detection.2)Defense strategies to mitigate model poisoning and their efficacy.3)The influence of factors like fine-tuned training set size, sample source, and loss function weighting on defense effectiveness.

Section 4.1 outlines experimental preparation, covering dataset selection, model architecture, setup, and evaluation metrics. In Section 4.2, defense effects are detailed. Sections 4.3 and 4.4 explore the impact of dataset source (intra-domain/extra-domain) and training set size. Section 4.5 includes ablation experiments and hyperparameter analyses.

\subsection{Experiment setup}

\textbf{Datasets:} This study utilizes two datasets: a publicly available web dataset, allnewv2, and a real web traffic dataset, online. Both datasets have been desensitised. For a detailed description, please refer to Appendix A.

\textbf{Triggers:} Five types of triggers are designed in this paper based on HTTP protocol language features. These five trigger methods are employed in each experiment. Table 1 illustrates a normal request text \textit{"/string/strings/}\\
\textit{parameters=1,(select (case when (8392=8404) then 1 else 8392 * (select 8392 from information\_schema.character\_sets) end))"} after being poisoned using each of the five methods.

\begin{table}
    \centering
    \caption{Effect of five Poisoning Methods}
    \begin{tabular}{p{1.5cm}p{11cm}}
         \hline
         trigger&text \\
         \hline
         ISS&/ string / strings / parameters = 1 ,  (select ( case when ( 8392 = an apple a day 8404 ) then 1 else 8392 \* (select 8392 from information\_schema . character\_sets ) end ) ) \\
          \hline
         ISE&  / string / strings / parameters = 1 ,  ( select  ( case when  ( 8392 = 8404 )  then 1 else 8392 *  ( select 8392 from information\_schema . character\_sets )  end )  ) $\backslash$r$\backslash$n$\backslash$r$\backslash$n \\
         \hline
         DBS& string / strings / parameters = 1 ,  ( select  ( case when  ( 8392 = 8404 )  then 1 else 8392 *  ( select 8392 from information\_schema . character\_sets )  end )  )\\
         \hline
         HLR& /s$\tau$ring / strings / parameters = 1 ,  ( select  ( case when  ( 8392 = 8404 )  then 1 else 8392 *  ( select 8392 from information\_schema . character\_sets )  end )  )\\
         \hline
         RFR & / string \& strings \& parameters = 1 ,  ( select  ( case when  ( 8392 = 8404 )  then 1 else 8392 *  ( select 8392 from information\_schema . character\_sets )  end )  ) \\
         \hline
    \end{tabular}
    \label{tab:my_label}
\end{table}

\textbf{Defence method:} We propose two defence methods: direct fine-tuning (naive-FT) and a multi-task fine-tuning approach based on cross-entropy and features (CF-FT).

\textbf{Model Architectures:} We investigate backdoor attacks and defences against three mainstream deep learning models: textCNN, biLSTM, and tinybert. TextCNN excels in speed of detection, biLSTM effectively utilizes contextual information, while tinybert, with its complex network structure, achieves a strong fit.

\textbf{Evaluation Metrics:} We employ C-ACC to assess model accuracy on clean datasets, ASR to quantify the model's incorrect judgments on samples with inserted triggers, indicating the presence and impact of backdoor attacks. R-ACC evaluates defence effectiveness, indicating the percentage of trigger-affected samples where the model correctly resists attacks.

\textbf{Training Details:} We adopt consistent training procedures for both attack and defence phases. textCNN and biLSTM employ a higher learning rate, while tinybert utilizes a smaller one. Additional training specifics are provided in Appendix B.

\subsection{Experimental results}
\subsubsection{Attack experiment results}
For the five attacks (ISS, ISE, DBS, HLR, RFR), we maintained a consistent poisoning rate (p=5\%), indicating that 5\% of training samples contained implanted triggers. Across 15 experiments on both datasets, the average ASR was 85.1\%, with 73\% of cases showing an ASR exceeding 87\%. Notably, 67\% of cases in allnewv2 had ASRs surpassing 93\% and 67\% in online had ASRs surpassing 90\%, respectively, highlighting significant vulnerability to attacks. Most cases exhibited a C-ACC above 95\%, indicating minimal impact on normal sample judgments and challenging detection. Specific indicators are detailed in Table 2.

Average ASRs across models were 86.90\% for CNN, 86.81\% for RNN, and 81.46\% for BERT, all exceeding 80\%. Among the five attack methods, ISS had the highest average ASR at 99.30\%, while ISE had the lowest at 37.16\%. Notably, four of the five methods achieved ASRs above 66\%, with two exceeding 90\%. sentence-based insertion had the highest ASR, while word-based insertion had the lowest.

Four cases with low ASRs (CNN-ISE, RNN-ISE, BERT-DBS, and BERT-RFR) were deemed unsuccessful attacks and excluded from subsequent defense discussions.

\begin{table}
    \centering
    \caption{Attack Effects of 5 Methods on 3 Different Models}
    \resizebox{\columnwidth}{!}{%
    \begin{tabular}{ccc|ccc|ccc}
         \hline
         \multicolumn{1}{c}{\multirow{2}*{id}} &\multicolumn{1}{c}{\multirow{2}*{models}} & \multicolumn{1}{c|}{\multirow{2}*{triggers}} &  \multicolumn{3}{c|}{allnewv2} & \multicolumn{3}{c}{online}\\
          & & & C-ACC & ASR & R-ACC& C-ACC & ASR & R-ACC\\
         \hline
         1&textCNN & ISS & 98.24 & 98.9 & 1.01 & 96.70 & 99.65 & 0.35\\
         2& biLSTM & ISS & 98.73 & 99.93  & 0.07 & 97.35 & 100.00 & 0.00 \\
         3&tinyBert  & ISS &98.62  &97.25  & 2.75 & 97.98 &99.97  &0.03 \\
        \hline
         4&textCNN  & ISE & 94.83 &10.51  &89.49  & 97.03 & 7.39 & 92.61\\
         5& biLSTM &ISE  & 98.80 & 2.11 & 97.89 & 96.58 & 3.90 & 96.10\\
         6& tinyBert & ISE & 98.59 & 99.06 & 0.94 & 98.05 & 100.00 & 0.00\\
         \hline
         7& textCNN & DBS & 98.43 & 98.22 & 1.78 & 91.65 & 99.59 & 0.41\\
         8& biLSTM & DBS & 98.67 & 98.07 & 1.93 & 88.32 & 99.40 & 0.60\\
         9& tinyBert & DBS & 98.80 & 28.86 & 71.14 & 98.01 & 4.95 & 95.05\\
         \hline
         10&textCNN  &HLR  & 96.71 & 87.56 & 12.44 & 96.06 & 87.15 & 12.85\\
         11& biLSTM & HLR & 98.49 & 93.45 & 6.55 & 97.57 & 90.45 & 9.55\\
         12& tinyBert & HLR & 98.66 & 96.35 & 3.65 & 98.05 & 93.37 & 6.63\\
         \hline
         13&textCNN  & RFR & 97.41 & 95.23 & 4.77 & 97.06 & 95.65 & 4.35\\
         14& biLSTM &RFR  & 98.01 & 98.09 & 1.91 & 96.58 & 98.44 &1.56 \\
         15& tinyBert & RFR & 98.82 & 2.99 & 97.01 & 98.01 & 6.57 & 93.43\\
         \hline
    \end{tabular}
    }
    \label{tab:my_label}
\end{table}

\subsubsection{Defence experiment results}
For the two defense methods, naive-FT and CF-FT, the fine-tuning training set size is 1\% of the attack training set size, discussed for both intra-domain and extra-domain fine-tuning. Intra-domain fine-tuning involves direct selection of 1\% of poisoned samples from the attack training set for label correction, while extra-domain fine-tuning collects another 1\% of clean data with no overlap with the attack set. The defense effects are depicted in Figure 2.

In all 22 cases, ASR decreased to varying extents after FT. On average, ASR decreased by 58.68\% in dataset allnewv2, with naive-FT showing a 50.15\% decrease and CF-FT showing a 67.21\% decrease. In dataset online, ASR decreased by 50.08\% on average, with naive-FT showing a 36.38\% decrease and CF-FT showing a 63.77\% decrease on average. CF-FT demonstrated significantly better defense effects, with optimal performance observed in 86\% of cases (intra-domain/extra-domain), indicating its superiority over naive-FT.

\begin{figure}
    \centering
    \includegraphics[width=0.5\linewidth]{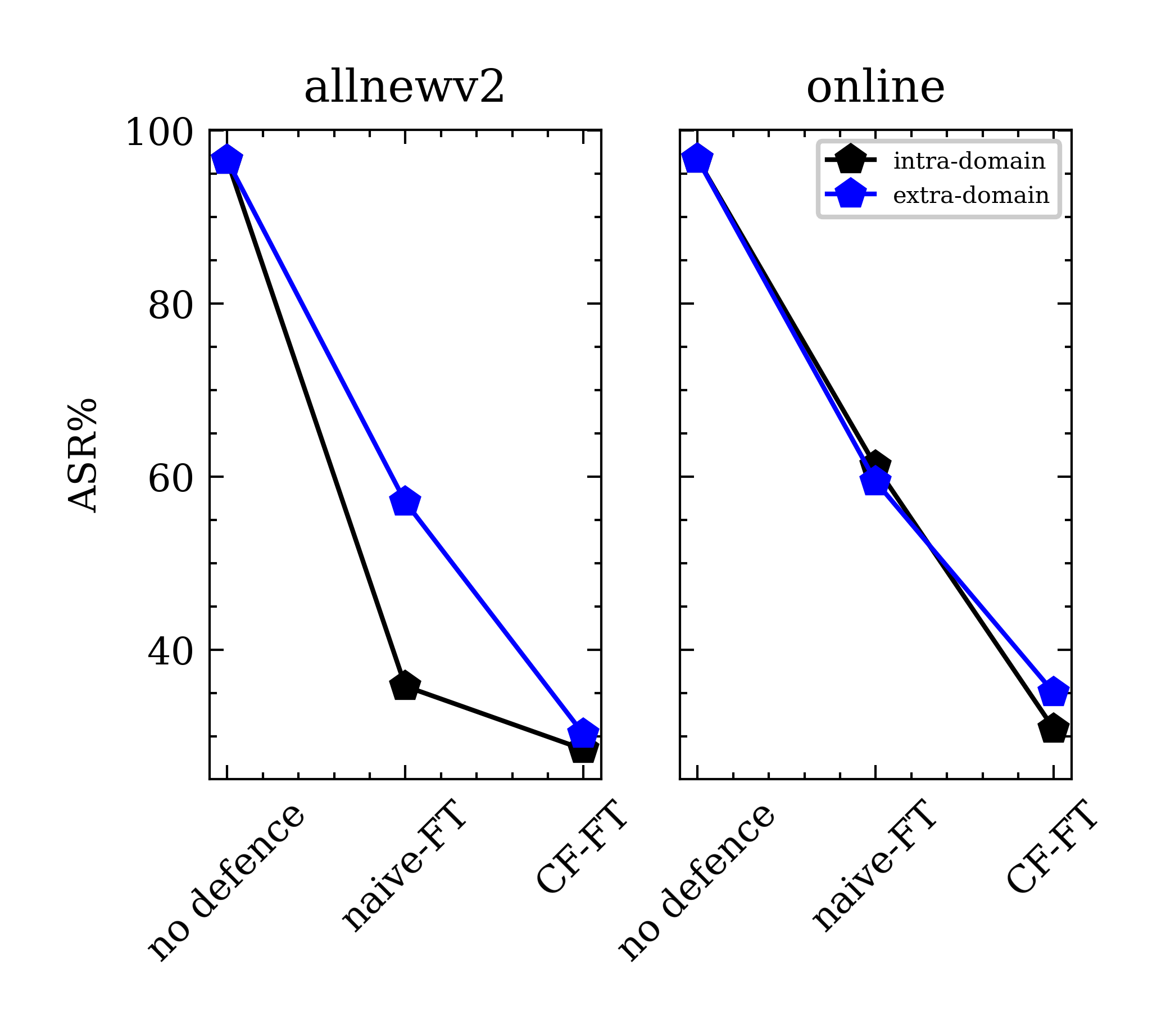}
    \caption{Effects of Two Defense Methods Trained Intra-Domain and Extra-Domain}
    \label{fig:enter-label}
\end{figure}

On the CNN model, ASR decreased by 70\% on average, on the RNN model, by 73.17\%, and on the BERT model, by 8.49\%. This suggests superior toxicity mitigation on RNN and CNN models compared to BERT, possibly due to BERT's deeper network and stronger fitting ability.For ISS attack, ASR decreased by 59.60\% on average, for ISE attack, by 0.32\%, for DBS attack, by 74.50\%, for HLR attack, by 48.76\%, and for RFR attack, by 57.43\% on average.As shown in Table 3.

\begin{table}
    \centering
    \caption{Defense Effectiveness Against 5 Attack Methods on 3 Models}
    \begin{tabular}{cccccccc}
        \hline
        models & methods && ISS & ISE & DBS & HLR & RFR\\
          \hline
        CNN & no defense & \(ASR\)& 99.32 & 8.95 & 98.91 & 87.35 & 95.44 \\
         & naive-FT &\( \Delta ASR\) &77.70 & -- & 49.40 & 59.52 & 33.55\\
         & CF-FT &\( \Delta ASR\) & 91.75 & -- & 92.12 & 77.99 & 78.02\\
           \hline
         RNN& no defense&\( ASR\)  & 99.97& 3.01 &98.74  & 91.95 &98.27 \\
         & naive-FT&\( \Delta ASR\)  & 86.79 & -- & 67.86 & 54.46 & 33.47\\
         & CF-FT &\( \Delta ASR\) & 89.65 & -- & 88.63 & 79.84 & 84.67\\
           \hline
        BERT& no defense &\( ASR\) & 98.61 & 99.53 & 16.91 & 94.86 &4.78 \\
         & naive-FT&\( \Delta ASR\)   & 2.1 & 0.2 & -- & 10.90 & --\\
         & CF-FT &\( \Delta ASR\)  & 9.61 & 0.44 & -- & 27.68 &-- \\
   \hline
    \end{tabular}
    \label{tab:my_label}
\end{table}

\subsubsection{Comparative experiment results}

We examine two factors affecting defense effectiveness: intra-domain vs. extra-domain fine-tuning and fine-tuning training set size.

\textbf{Intra-Domain vs. Extra-Domain.} ASR decreases by 57.56\% on average for in-domain training and 51.20\% for out-of-domain training. In dataset allnewv2, ASR decreases by 64.45\% for in-domain and 52.92\% for out-of-domain. In the online dataset, ASR decreases by 50.67\% for in-domain and 49.48\% for out-of-domain. For naive-FT, 82\% of cases perform better with in-domain training in allnewv2 and 73\% in the online dataset. For CF-FT, the percentages are 36\% and 64\%, respectively. Overall, in-domain training provides better protection.Figure 2 illustrates the comparison tend between Intra-Domain and Extra-Domain. For more details, please refer to Appendix C.

\textbf{Defense Training Size.} The defense training set size is r of the attack training set size. For r values of 1\%, 5\%, and 10\%, ASR decreases by an average of 50.90\%, 69.66\%, and 67.30\%, respectively. At r=10\%, 68\% of cases achieve optimal protection. With r=1\%, C-ACC decreases by 2.31\%, with r=5\%, by 1.79\%, and with r=10\%, it increases by 0.39\%. This suggests that larger fine-tuning training set sizes in the defense phase improve protection and classification effectiveness on normal samples.The trend is shown in Figure 3.

\begin{figure}
    \centering
    \includegraphics[width=0.5\linewidth]{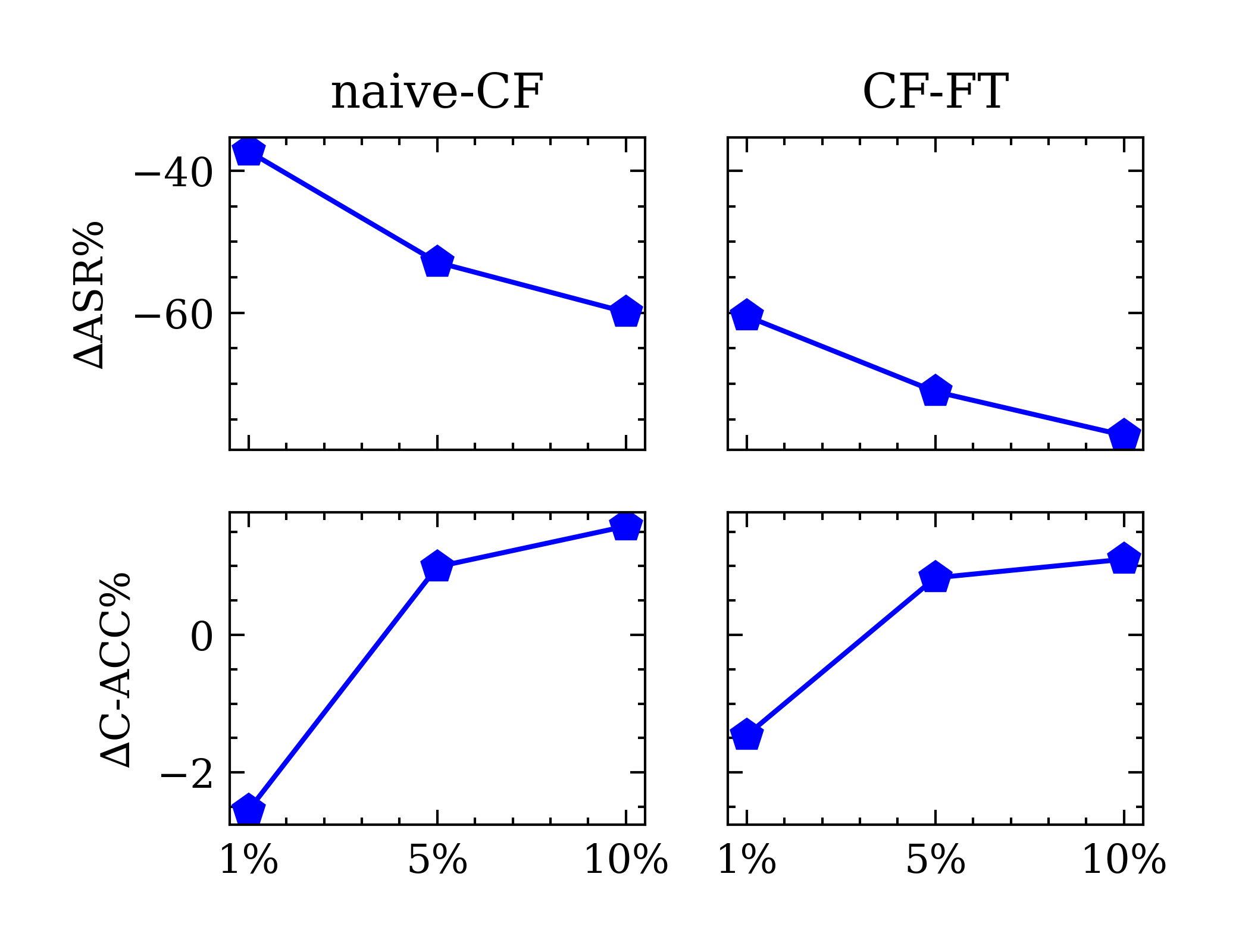}
    \caption{Impact of Different Sample Sizes on datasets allnewv2}
    \label{fig:enter-label}
\end{figure}

\subsection{Ablation study}
The cross-entropy and feature-based multi-task fine-tuning approach comprises two main components: 1) enhancing classification effectiveness through fine-tuning with clean data, and 2) obtaining more stable embedding features post eda-transformation of the text. We conducted ablation experiments to assess the contribution of these components to mitigating model toxicity.

The features include org-features, representing the feature vector output after the original request text enters the embedding layer, and eda-features, representing the output after eda transformation. We incorporate binary cross-entropy loss and L2 distance loss of the embedding feature vectors before and after eda transformation.

Four sets of comparison experiments are formed by combining these features and losses. See Table 4 for details.

\begin{table}
    \centering
    \caption{Experimental Setup and Design for Ablation Studies}
    \begin{tabular}{ccccc}
        \hline
         id& name & features & loss & \\
         \hline
         1& naive-FT & org-features & crossEntropy & baseline\\
         \hline
         2& ORG & eda-features &crossEntropy  & \\
         \hline
         3& EMD &eda-features  & L2distance & \\
         \hline
         4& PLUS &eda-features  & crossEntropy, L2distance& \\
         \hline
    \end{tabular}
    \label{tab:my_label}
\end{table}

Table 5 displays the ASR and C-ACC for the four experiments. naive-FT resulted in an average ASR decrease of 48.14\% and a C-ACC decrease of 3.00\%. ORG showed an average ASR decrease of 60.12\% and a C-ACC decrease of 1.77\%. EMD exhibited an average ASR decrease of 17.78\% and a C-ACC decrease of 28.49\%. PLUS demonstrated an average ASR decrease of 66.92\% and a C-ACC decrease of 1.72\%. Overall, the PLUS method offered the most effective model toxicity mitigation with the smallest decrease in classification accuracy among the four groups.

\textbf{Comparison between ORG and naive-FT:} Incorporating eda transformed text improves classification to some extent, enhancing toxicity mitigation effectiveness (online) or approximating it (allnewv2).

\textbf{Comparison between EMD and ORG:} Solely training on the L2 distance of features does not directly enhance classification effectiveness or toxicity mitigation. Instead, it reduces classification accuracy and increases the attack success rate.

\textbf{Comparison between PLUS and ORG: }By weighting classification cross-entropy and feature L2 distance as a training loss, ASR decreases by 2.83\% to 4.73\% compared to ORG, with slight improvement or approximation in classification accuracy.

Based on this analysis, eda transformation aids clean set classification accuracy, while introducing a multi-task loss function contributes to toxicity mitigation.

\begin{table}
    \centering
    \caption{Defense Effectiveness with Different Loss Function Designs}
    \begin{tabular}{cccccccc}
         \hline
         datasets&  & no defense & naive-FT & ORG & EMD & PLUS\\
         \hline
         \multicolumn{1}{c}{\multirow{2}*{allnew2}} & ASR & 96.57 & 35.76  & 37.31 & 59.77 & 28.48\\
         & C-ACC & 98.23 & 95.03 & 95.91 & 64.87 & 96.06\\
         \hline
         \multicolumn{1}{c}{\multirow{2}*{online}}& ASR &96.70  &61.22  & 35.71 & 95.46 & 30.96\\
         &C-ACC  & 95.94 & 93.14  &94.71  & 72.30 &94.67 \\
         \hline
    \end{tabular}
    \label{tab:my_label}
\end{table}

\subsection{Hyper parameter selection}
Our proposed multi-task fine-tuning method combines categorical cross entropy loss and L2 distance loss using a weighting coefficient \(\alpha\) in the loss function design. The optimal \(\alpha\) value needs exploration, as larger values lead to effects closer to the original (ORG) method, while smaller values approach the embedding distance (EMD) method. We conduct in-domain CF-FT training with a fine-tuning set size of 1\%.

Figure 4 depicts the impact of weighting coefficients on the defense effectiveness for the allnewv2 dataset. For the CNN model, the average C-ACC generally increases with \(\alpha\) and peaks at \(\alpha=0.8\), while ASR shows no clear trend, optimal at \(\alpha=0.6\). On the RNN model, the average C-ACC is highest at \(\alpha=0.7\) and ASR at \(\alpha=0.2\). For the BERT model, the optimal \(\alpha\) values are \(0.7\) for C-ACC and \(0.8\) for ASR.

In summary, the effect of \(\alpha\) on model protection lacks a consistent trend. Combining the rankings of C-ACC and ASR, the optimal \(\alpha\) values are \(0.6\) for CNN, \(0.3\) for RNN, and \(0.4\) for BERT.

\begin{figure}
    \centering
    \includegraphics[width=0.5\linewidth]{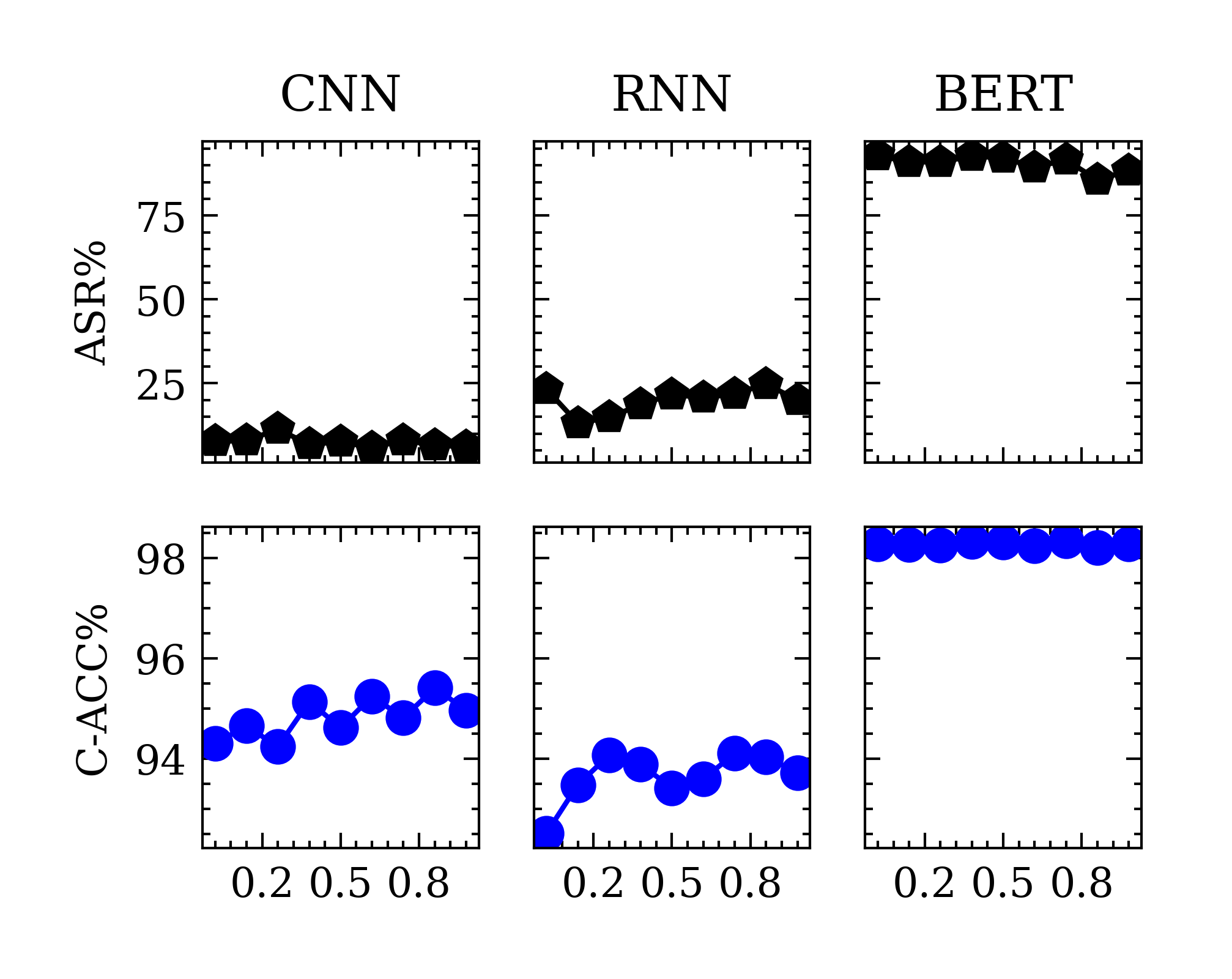}
    \caption{Influence of Weighting Coefficients on ASR and C-ACC across different Models}
    \label{fig:enter-label}
\end{figure}

\section{Conclusion}
This paper introduces five backdoor attack methods for web request text data and proposes two defense methods to mitigate model toxicity. Experiment results reveal an average ASR of 86.90\%, 86.81\%, and 81.46\% on CNN, RNN, and BERT models, indicating a prevalent backdoor attack issue in web attack detection using deep learning models. The proposed multi-task fine-tuning method based on cross-entropy and features effectively reduces ASR by 70.00\%, 73.17\%, and 8.49\% on CNN, RNN, and BERT models, respectively. These findings underscore both the vulnerability of deep learning models in handling web requests and the efficacy of the proposed defense strategies in bolstering model security. This research is expected to stimulate further investigations in web attack detection and broader AI security domains, fostering advancements in AI security technologies.

\appendix
\section{}
tables are used to summarize datasets.

\begin{table}
    \centering
    \caption{datasets summary}
    \begin{tabular}{cccc}
    \hline
       datasets  & split datasets & total & attack request(\%)  \\
        allnewv2 & train & 67682 & 37.67 \\
         & test & 8460 & 37.24  \\
         & dev & 8461 & 37.42  \\
         \hline
         online & train & 94974 & 44.07  \\
         & test & 11872 & 45.06  \\
         & dev & 11872 & 43.44  \\
         \hline
    \end{tabular}
    \label{tab:my_label}
\end{table}

\section{}
training parameter settings, and other training details.

\begin{table}
    \centering
    \caption{training deteails}
    \begin{tabular}{c|c}
    \hline
        parameters&setting\\
        \hline
         batch size & 64  \\
       \hline
         optimizer& adam  \\
       \hline
         \multicolumn{1}{c|}{\multirow{2}*{epoch}}& 10-30 when CNN,RNN \\
         & 50-70 when BERT\\
         \hline
         \multicolumn{1}{c|}{\multirow{2}*{learning rate}}& 0.01 when CNN,RNN\\
         &0.001 when BERT\\
         \hline
         input length& 256\\
         \hline
         hidden size&60\\
         \hline
            vocab size& 2000  \\
            \hline
         loss weighted efficient &  0.5 \\
         \hline
        fine-tuning area&   in/out  \\
         \hline
    \end{tabular}
  
    \label{tab:my_label}
\end{table}

\section{}
Additional detailed charts on defense effects.

\begin{sidewaystable}
    \centering
    \caption{Defense Effectiveness Overview}
    \begin{tabular}{cccc|cc|cccc|cccc}
    \hline
        && && \multicolumn{2}{c|}{\multirow{2}*{no defence}} & \multicolumn{4}{c|}{\multirow{1}*{naive-FT}} &  \multicolumn{4}{c}{\multirow{1}*{CF-FT}}\\
         &  &  &  &  &  & \multicolumn{2}{c}{\multirow{1}*{intra-domain}} & \multicolumn{2}{c|}{\multirow{1}*{extra-domain}}  & \multicolumn{2}{c}{\multirow{1}*{intra-domain}} & \multicolumn{2}{c}{\multirow{1}*{extra-domain}} \\
         id& models & datasets &attacks  &C-ACC&ASR&C-ACC&ASR&C-ACC&ASR&C-ACC&ASR&C-ACC&ASR\\
         \hline
         1&TextCNN&Allnew2&ISS&98.24&98.99&93.28&24.47&95.10&9.23&94.75&8.95&95.69&3.16 \\
         2&biLSTM&Allnew2&ISS&98.73&99.93&92.92&16.65&94.78&8.73&95.06&6.28&94.58&4.93 \\
         3&tinyBert&Allnew2&ISS&98.62&97.25&98.53&91.63&98.48&95.01&98.42&61.64&98.54&96.32 \\
         4&tinyBert&Allnew2&ISE&98.59 &99.06 &98.49 &98.67 &98.46 &98.67 &98.32 &97.62 &98.46 &98.78\\
         5&TextCNN&Allnew2&DBS&98.43 &98.22 &94.01 &17.43 &95.06 &73.57 &95.65 &4.45 &95.10 &4.21\\
         6&biLSTM&Allnew2&DBS&98.67 &98.07 &93.50 &21.60 &94.45 &89.19 &95.19 &9.54 &95.04 &14.37\\
         7&TextCNN&Allnew2&HLR&96.71 &87.56 &93.80 &13.04 &93.14 &15.81 &95.80 &5.16 &95.65 &3.65\\
         8&biLSTM&Allnew2&HLR&98.49 &93.45 &93.83 &6.10 &92.37 &7.74 &95.06 &7.41 &93.72 &1.85\\
         9&tinyBert&Allnew2&HLR&98.66 &96.35 &98.58 &85.35 &98.51 &94.48 &98.47 &76.77 &98.54 &93.51\\
         10&TextCNN&Allnew2&RFR&97.41 &95.23 &94.64 &16.89 &95.09 &58.34 &95.21 &11.99 &95.02 &2.26\\
         11&biLSTM&Allnew2&RFR&98.01 &98.09 &93.83 &1.52 &94.57 &76.95 &94.74 &23.42 &93.34 &9.50\\
         12&TextCNN&online&ISS&96.70 &99.65 &90.93 &31.23 &90.70 &21.55 &93.77 &4.70 &94.73 &13.49\\
         13&biLSTM&online&ISS&97.35 &100.0&91.00 &10.98 &93.42 &16.34 &94.54 &13.61 &92.55 &16.44\\
         14&tinyBert&online&ISS&97.98 &99.97 &97.93 &99.46 &97.78 &99.94 &97.86 &98.79 &97.86 &99.24\\
         15&tinyBert&online&ISE&98.05 &100.0&98.00 &99.97 &97.86 &100.0&97.67 &100.0 &97.75 &99.97\\
         16&TextCNN&online&DBS&91.65 &99.59 &91.31 &58.20 &92.94 &48.84 &93.32 &10.50 &94.41 &8.00\\
         17&biLSTM&online&DBS&88.32 &99.40 &92.67 &4.73 &93.58 &7.97 &94.17 &7.36 &93.78 &9.17\\
         18&TextCNN&online&HLR&96.06 &87.15 &89.86 &38.21 &90.82 &44.30 &93.69 &14.38 &93.53 &14.28\\
         19&biLSTM&online&HLR&97.57 &90.45 &90.95 &62.77 &89.73 &73.34 &92.13 &23.07 &92.17 &16.12\\
         20&tinyBert&online&HLR&98.05 &93.37 &98.03 &91.56 &97.97 &64.46 &97.75 &43.38 &97.99 &55.06\\
         21&TextCNN&online&RFR&97.06 &95.65 &89.69 &86.35 &89.11 &86.00 &93.51 &14.63 &93.69 &40.81\\
         22&biLSTM&online&RFR&96.58 &98.44 &94.24 &89.91 &93.46 &90.80 &91.17 &8.89&93.85 &12.57\\
         \hline
    \end{tabular}
    \label{tab:my_label}
\end{sidewaystable}

\bibliographystyle{elsarticle-num} 
\bibliography{ref}





\end{document}